\documentclass[times,twocolumn,final,authoryear]{elsarticle}
\usepackage{lipsum}
\usepackage{amsmath}
\usepackage{amssymb}
\usepackage{times}
\usepackage{epsfig}
\usepackage{graphicx}
\usepackage{amsmath}
\usepackage{amssymb}
\usepackage{tabularx}
\usepackage{multirow}
\usepackage{subfigure}
\usepackage{comment}
\usepackage{float}
\usepackage{wrapfig}
\usepackage[ruled,vlined]{algorithm2e}
\usepackage{algorithmic}
\usepackage{bm}
\usepackage{color}

\usepackage{prletters}
\usepackage{framed,multirow}

\usepackage{amssymb}
\usepackage{latexsym}

\usepackage{url}
\usepackage{xcolor}

\usepackage{xspace}
\makeatletter
\DeclareRobustCommand\onedot{\futurelet\@let@token\@onedot}
\def\@onedot{\ifx\@let@token.\else.\null\fi\xspace}

\def\etal{et al\onedot}
\usepackage{booktabs}

\definecolor{newcolor}{rgb}{.8,.349,.1}
\journal{Pattern Recognition Letters}
\begin{document}
\begin{frontmatter}
\title{Asymmetric kernel in Gaussian Processes for learning target variance}
\author[1]{S. L. \snm{Pintea} 
\corref{cor1}} 
\ead{S.L.Pintea@tudelft.nl}
\author[1]{J. C. \snm{van Gemert}}
\author[2]{A. W. M. \snm{Smeulders}}
\address[1]{
	Computer Vision Lab, Delft University of Technology, Delft, Netherlands\\
}
\address[2]{
	Intelligent Sensory Information Systems, University of Amsterdam, Amsterdam, Netherlands\\
}

\received{28 March 2017}
\finalform{28 March 2017}
\accepted{28 March 2017}
\availableonline{28 March 2017}
\communicated{S. Sarkar}

\begin{abstract}
This work incorporates the multi-modality of the data distribution into a Gaussian Process regression model.
We approach the problem from a discriminative perspective by learning, jointly over the training data, 
the target space variance in the neighborhood of a certain sample through metric learning.
We start by using data centers rather than all training samples. 
Subsequently, each center selects an individualized kernel metric. 
This enables each center to adjust the kernel space in its vicinity in correspondence with the topology
of the targets --- a multi-modal approach.
We additionally add descriptiveness by allowing each center to learn a precision matrix.
We demonstrate empirically the reliability of the model.    
\end{abstract}


\end{frontmatter}
\section{Introduction}
\label{sec:intro}
\noindent Departing from the standard Gaussian Process, we introduce a regression approach that incorporates the multi-modality of the data distribution.
While in the Gaussian Process model we have a global kernel metric that is shared by all the samples \cite{rasmussen2006gaussian},
here we propose to define a set of training data centers considerably smaller than the number of training samples.
Subsequently, we learn from the numerous training samples an individualized kernel metric per training data center.
By doing so, we are able to use a smaller training kernel matrix computed only on the training data centers while 
retaining the descriptive power of the model. 
This is highly efficient at test-time as it limits the size of the kernel matrix.

\begin{figure*}
	\centering
	\rotatebox{-90}{\includegraphics[width=0.33\textwidth]{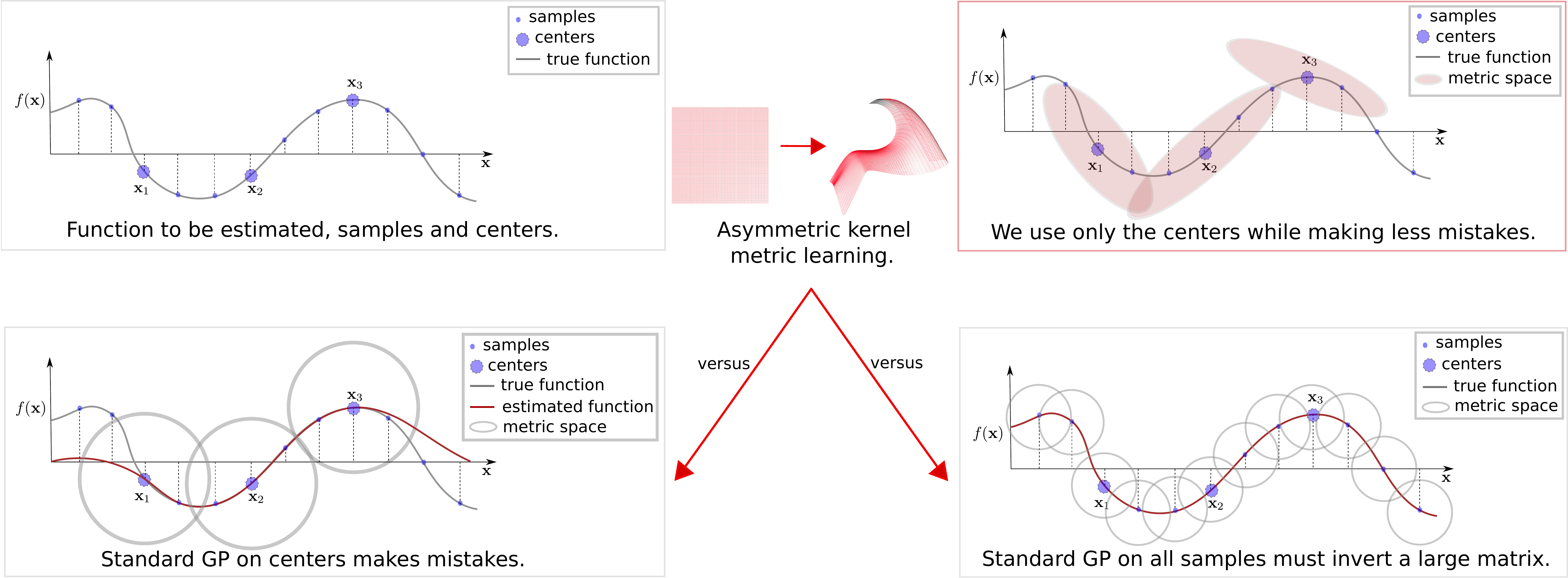}}
	\caption{\small An intuitive illustration of the proposed asymmetric kernel metric optimization, when compared with the standard Gaussian Process on data centers or on all samples. 
		Given that each data center learns both a personalized size and shape of the kernel, we obtain a more descriptive model than the standard Gaussian Process, 
		while using a limited number of data centers in the kernel matrix computation.    
		}
	\label{fig:abstract}
\end{figure*}
In our method we introduce two main changes to the standard Gaussian Process regressor:
(i) we define a number of centers over the training data, by clustering or sampling; 
(ii) we learn individual kernel metric parameters per data center, discriminatively through metric learning, giving rise to a multi-modal approach with an asymmetric kernel matrix.
Figure~\ref{fig:abstract} illustrates these differences when comparing with the standard Gaussian Process trained on either data centers or on all samples: 
we use fewer samples in the kernel computation, while enhancing the descriptiveness of the model by learning individualized metrics per center.
We additionally expand the kernel parameters to a precision matrix, also learned through metric learning per center, giving rise to a multivariate multi-modal approach. 

The individual steps of our model are not specifically novel, yet their combination is what gives strength. 
Clustering of the data in Gaussian Processes has been previously proposed \cite{mycviu,snelson2007local}. 
Asymmetric kernels have been studied in works such as \cite{mackenzie2004asymmetric, wua2010asymmetric}.
While a multivariate lengthscale parameter has been used in \cite{kersting2007most,lazaro2011variational,mycviu} for improved descriptiveness. 
Here, we combine all these ideas into a new approach which is suitable for learning target data variance.
If we consider each training sample to be a data center, and enforce that all samples share the same kernel metric, 
and assume a univariate lengthscale in the kernel metric, we recover the standard Gaussian Process definition.
We evaluate the proposed approach by gradually enabling these changes: center-based Gaussian Process, univariate multi-modal asymmetric Gaussian Process, 
and multivariate multi-modal asymmetric Gaussian Process.
The experiments validate our models on the regression datasets \emph{So2} and \emph{Temp} 
used in \cite{titsias2013variational}, the large scale \emph{Airlines} dataset of \cite{hoang2015unifying}, and two realistic image datasets: UCSD \cite{chan2012counting}
and VOC-2007 \cite{pascal-voc-2007} for pedestrian and generic-object counting, respectively.
 
\section{Related Work}
\subsection{Mixtures of Gaussian Processes}
\noindent Noteworthy work has been focusing on mixtures of Gaussian Processes 
\cite{lazaro2012overlapping,meeds2006alternative,nguyen2014fast,rasmussen2002infinite,tresp2000mixtures,yuan2009variational}.
In \cite{tresp2000mixtures} a mixture of Gaussian Processes is proposed to effectively deal with large data.
\cite{meeds2006alternative,rasmussen2002infinite} extend this idea to an infinite mixture of Gaussian Processes.
Somewhat similarly, \cite{li2014gaussian} splits the problem into subproblems in a divide-and-conquer fashion and solves each such problem in a Gaussian Process model. 
\cite{yuan2009variational} proposes an elegant variational Bayesian algorithm for training the mixture of Gaussian Process experts.
An effective model is proposed in \cite{lazaro2012overlapping}, where the authors simplify the mixture of experts so no gating 
function is used to assign samples to components and rather, trajectory clustering is employed. 
\cite{nguyen2014fast} gracefully combines the mixture of Gaussian Process experts with the idea of inducing points, providing 
fast approximate Gaussian Process models.
Unlike these works, where the final prediction entails a combination of predictions, each obtained within the metric space of individual components, 
we learn the hyper-parameters associated with each training data centers in a single Gaussian Process.
We do so by employing an asymmetric kernel. 
Therefore, at test-time for an input test sample, rather than computing $N \times M$ kernel distances, where $N$ is the number of components and $M$ is the number of training samples, 
we only compute $N$ distances.
In our case $N$ is the number of data centers, considerably smaller than $M$. 

\subsection{Efficiency in Gaussian Processes}
\noindent The work of \cite{quinonero2005unifying} reviews the sparse approximations of Gaussian Process from a unified perspective by analyzing 
the implied prior of different methods.
\cite{csato2002sparse} proposes learning iteratively, online, the sparse set of inducing points in a Bayesian formalism by minimizing the KL divergence.
Inspired from metric learning techniques, \cite{lawrence2003fast} uses forward selection to obtain a sparse and time-efficient model. 
\cite{snelson2005sparse} proposes a graceful solution of learning a set of sparse pseudo-inputs through gradient based optimization.
A combination between sparse methods based on inducing points and local regression based on a multitude of experts 
describing locally the target space, is proposed in \cite{snelson2007local}.
In \cite{titsias2009variational,titsias2010bayesian} variational approaches are used to learn sparse representations.
\cite{titsias2009variational} jointly learns the inducing points and the kernel hyper-parameters by minimizing a lower bound 
through KL divergence.
The robust method of \cite{hensman2013gaussian} decomposes the Gaussian Process model, variationally, such that it is 
factorized based on a set of global inducing variables.  
\cite{bo2012greedy,ranganathan2011online} focus on iterative updates of the Gaussian Process. 
\cite{rodner2012large} proposes the use of parameterized histogram intersection kernels to bypass the hyper-parameter 
estimation. 
\cite{cao2013efficient} proposes a method to speed up the hyper-parameter estimation by inducing sparsity in the model. 
Somewhat similar to these methods, we only retain a set of data centers as informative training samples.
Yet, unlike the above approaches, we subsequently add extra information into the Gaussian Process model by treating the data centers differently. 

\subsection{Descriptiveness in Gaussian Processes}
\noindent Full matrices in the kernel definition have been proposed in \cite{mycviu,vivarelli1999discovering}, to make the model more descriptive.
Here, we also learn precision matrices, in the kernel metric definition.
However in our work, each center has an individualized precision matrix.
\cite{paciorek2004nonstationary} proposes nonstationary covariance matrices in the Gaussian Process model, tying the kernel metrics to the input samples. 
However, the final kernel matrix is symmetric as it is defined using symmetric combinations of per-sample covariances, similar to RVM (Relevance Vector Machine).
\cite{kersting2007most,lazaro2011variational} propose well-founded approaches to adding descriptiveness by extending the Gaussian Process definition to a 
heteroscedastic approach, by modeling the noise distribution to be dependent on the training data.
\cite{kuss2005approximate} proposes EP (Expectation Propagation) as an effective manner to train these models.
Similarly, we also start with the assumption that the kernel metric should be data dependent and learn an individualized kernel metric. 
\cite{titsias2013variational} proposes a compelling method for adjusting the kernel distances by assuming the data is mapped in a feature space based on the 
Mahalanobis kernel distance, estimated through variational inference. 
Unlike this work, we learn both the shape and the scale of the kernels per data center by minimizing the predictive loss.

\subsection{Asymmetric Kernels}
\noindent The use of asymmetric kernel distances is not a recent idea but, rather, a well-matured topic \cite{kulis2011you, mackenzie2004asymmetric, tsuda1999support, wua2010asymmetric}.
In \cite{tsuda1999support} asymmetric kernels are proposed in the context of SVM classification.
\cite{wua2010asymmetric} shows how similarity functions commonly used in real-life applications, can be related to asymmetric kernels, 
and gives a formal definition for the mathematical space described by asymmetric kernels.
The work in \cite{mackenzie2004asymmetric} proposes asymmetric kernel regression in the context of neural networks and shows that such models
are better behaved around the data boundaries. 
The recent work of \cite{kulis2011you} learns asymmetric distances for visual domain adaptation in the context of object recognition. 
Similar to these methods, we also use asymmetric kernel distances, as these prove to have more descriptive power when 
limiting the number of samples in the training kernel computation.
 
\subsection{Metric Learning}
\noindent \cite{rahimi2007random} learns a lower dimensional mapping of data while maintaining the distances between 
the data samples --- the kernel distances remain approximately equal to the ones of the original features. 
In our work, we learn the kernel metric given the targets, rather than the feature representation.
In \cite{weinberger2007metric} the kernel metric minimizes the leave-one-out regression error. 
\cite{jain2012metric} combines kernel learning with metric learning by employing a linear transformation.
In this work, we use a fixed kernel --- the squared exponential kernel, we additionally expand the parameters of 
the kernel to full precision matrices.
\cite{globerson2005metric,weinberger2009distance,xing2002distance} represent pioneering work in the field of metric learning. 
\cite{xing2002distance} is the first paper to pose metric learning as a convex optimization problem learned from 
similar\slash dissimilar pairs of points.
\cite{globerson2005metric} is one of the first works to propose Mahalanobis distances for metric learning.
In \cite{weinberger2009distance} the Mahalanobis distance is learned in a nearest neighbor classifier, which induces a large-margin separation of classes.
\cite{kedem2012non,kostinger2012large,weinberger2009distance} are recent works focusing on metric learning for classification with kernels, 
while \cite{huang2013kernel} focuses on sparse kernel learning for regression. 
In this work we employ metric learning rather than estimating the optimal model hyper-parameters through marginal likelihood 
\cite{rasmussen2006gaussian}.
We do so, as each data center has an associated lengthscale in the proposed model and, thus, the marginal 
likelihood optimization is not straightforward in our case. 

\section{Asymmetric Kernel for Gaussian Processes}
\noindent We redefine the Gaussian Process model by allowing each training data center to learn an individualized kernel metric.
This entails that the kernel matrix ceases to be symmetric in our case.
However, this comes at extra gain in descriptive power, as despite using a small set of samples in the training kernel matrix,
we optimize the individualized kernel metrics over the numerous available training samples.

\subsection{Standard Gaussian Process Revisited}
\noindent We shortly revisit the standard Gaussian Process formulation, to unify the notations.
The mean of the predictive distribution is \cite{rasmussen2006gaussian}:
\begin{alignat}{1}
	f(\mathbf{x}^*) &= k(\mathbf{X},\mathbf{x}^*)^T \left( k(\mathbf{X},\mathbf{X}) + \sigma^2\mathbf{I} 
		\right)^{-1} \mathbf{y},
		\label{eq:inverse}
\end{alignat}
Where $\mathbf{x}^*$ represents an input test sample, $\mathbf{X}$ represents the training samples used for the training kernel matrix computation,
$\mathbf{y}$ represents the training targets, $f(\mathbf{x}^*)$ is the prediction over the input $\mathbf{x}^*$ and $k(\cdot, \cdot)$ is the kernel 
metric used for estimating sample distances, and $\sigma$ is the noise hyper-parameter.

\subsection{Center-based Gaussian Process}
\noindent As equation \ref{eq:inverse} indicates, the training procedure requires the computation of the inverse of the training kernel-matrix, 
$k(\mathbf{X},\mathbf{X})$, which is prohibitive on larger datasets.
The first alteration of the Gaussian Process model that we investigate, is considering a set of data centers rather than individual training samples.
We do so by either sampling the data or clustering it into a set of centers, $\overline{\mathbf{X}}$.
Despite its simplicity, this is very effective in getting a fair overview over the variation in the training data while not having to use 
all samples during training.  

More principled manners of defining data centers such as effective sampling techniques are possible. 
However, the focus here is not on the center definition, which is just meant as a first step towards reducing the size of the training kernel matrix.
The strength of our model comes from allowing these centers to learn individualized metrics.     

\subsection{Multi-modal Asymmetric Kernel}
\noindent Given that we have sparsified the training data by keeping only the training data centers, we lost information regarding the smoothness or 
variability of the target function in different regions of the data space.
Therefore, we allow each training center, $\overline{\mathbf{x}}_i \in \overline{\mathbf{X}}$, to define individualized kernel metrics in its data neighborhood.
The lengthscale hyper-parameter is the one defining the size of the kernel space, thus, we propose individualized lengthscale hyper-parameters, $l_i$ for each center $\overline{\mathbf{x}}_i$.
This entails the second alteration of the standard Gaussian Process model.
In this case the prediction function uses training and test kernel terms with per-center metrics:
\begin{alignat}{1}
	f(\mathbf{x}^*) &= \hat{k}(\overline{\mathbf{X}},\mathbf{x}^*)^T \left( \hat{k}(\overline{\mathbf{X}},\overline{\mathbf{X}})+
		\sigma^2I \right)^{-1} \mathbf{y},
	\label{eq:pred_center}
\end{alignat}
where $\hat{k}(\cdot,\cdot)$ is a non-symmetric kernel whose size depends on its corresponding 
training center --- $\hat{k}(\overline{\mathbf{x}}_i,\overline{\mathbf{x}}_j) = k_i (\overline{\mathbf{x}}_i,\overline{\mathbf{x}}_j)$, and 
$\overline{\mathbf{x}}_i, \overline{\mathbf{x}}_j \in \overline{\mathbf{X}}$ are data centers.
Thus, the distance from a training center to the others is computed within the associated kernel space of that center.
At test-time $\hat{k}(\overline{\mathbf{X}},\mathbf{x}^*) = \left(k_1(\overline{\mathbf{x}}_1,\mathbf{x}^*), k_2(\overline{\mathbf{x}}_2,\mathbf{x}^*),.. 
k_N(\overline{\mathbf{x}}_N,\mathbf{x}^*)\right)$, where $N$ is the number of training centers, and $\mathbf{x}^*$ is a test sample. 
In this work we restrain our focus to the squared exponential kernel distance:
\begin{alignat}{1}
	\hat{k}(\overline{\mathbf{x}}_i,\overline{\mathbf{x}}_j) = k_i(\overline{\mathbf{x}}_i,\overline{\mathbf{x}}_j) &= \text{exp}\left( -\frac{1}{2 l^2_i} 
		(\overline{\mathbf{x}}_i-\overline{\mathbf{x}}_j)(\overline{\mathbf{x}}_i-\overline{\mathbf{x}}_j)^T \right).
	\label{eq:univariate}
\end{alignat}
where $l_i$ is the lengthscale associated with data center $\overline{\mathbf{x}}_i$.

Given the use of individualized metrics per data center, the kernel ceases to be symmetric.
Therefore, we can no longer employ the standard Cholesky decomposition for estimating the kernel-matrix inverse.
We compute the kernel matrix inversion through SVD (Singular Value Decomposition).  
Despite this drawback, the individualized kernel metrics allow us to optimize the scale of the kernel 
locally, in the neighborhood of each data center.   

\subsection{Multivariate Multi-modal Asymmetric Kernel}
\noindent By allowing each center to define its own kernel metric, we change the model such that we can locally resize the kernel space
to better map the target space. 
As highlighted in \cite{chen2013stress}, not only the size of the kernel space is important but also the shape. 
Therefore, we also consider a multivariate extension of the model that allows for optimizing also the kernel shape per training center.  
\begin{alignat}{2}
	\hat{k}(\overline{\mathbf{x}}_i,\overline{\mathbf{x}}_j) = k_i(\overline{\mathbf{x}}_i,\overline{\mathbf{x}}_j) &= \text{exp}\left( -\frac{1}{2} 
		(\overline{\mathbf{x}}_i-\overline{\mathbf{x}}_j) \bm{P}_i (\overline{\mathbf{x}}_i-\overline{\mathbf{x}}_j)^T \right).
	\label{eq:multivariate}
\end{alignat}
where $\bm{P}_i$ is the precision matrix associated with the data center $\overline{\mathbf{x}}_i$, to be learned from training data 
samples other than the ones defining data centers.

\subsection{Kernel Metric Optimization}
\label{ssec:kernel}
\noindent Standardly in the literature, the Gaussian Process hyper-parameters are learned through gradient methods by maximizing the marginal 
likelihood over the weights.
Given that we aim to optimize a kernel metric, following the metric learning literature \cite{chen2013stress,globerson2005metric,weinberger2009distance,xing2002distance} 
we approach this problem discriminatively, and optimize the hyper-parameters with respect to the squared loss. 

Thus, we learn the appropriate kernel metric for each data center, $\overline{\mathbf{x}}_i \in \overline{\mathbf{X}}$, 
from training samples other than the data centers, $\mathbf{x}_n \in \mathbf{X}\setminus \overline{\mathbf{X}}$. 
We add a regularization term to the squared loss weighted by $\mu$.
We use the regularized squared loss over the targets as the function to be minimized and we employ SGD (Stochastic Gradient Descent) 
by estimating the gradients with respect to each per-center lengthscale, $l_i$ in the univariate case and $\mathbf{P}_i$ in the multivariate case.
\begin{alignat}{1}
	\mathcal{L}(f, \mathbf{y}^*) &= 
			\left\{
                \begin{array}{ll} 
			\sum_{i=1}^N \left( \sum_{\mathbf{x}_n \in \mathbf{X} \setminus \overline{\mathbf{X}}} ( f(\mathbf{x}_n) - y_n^*)^2 
			+  \mu l_i^2 \right), & \text{\scriptsize if univariate;}\\
			\sum_{i=1}^N \left( \sum_{\mathbf{x}_n \in \mathbf{X} \setminus \overline{\mathbf{X}}} ( f(\mathbf{x}_n) - y_n^*)^2 
			+  \mu \mid\mid \bm{P}_i \mid\mid \right), & \text{\scriptsize multivariate.}
			\end{array}
        \right.
	\label{eq:loss}
\end{alignat}	
We denote by $\mathbf{y}^*$ the training target vector composed of values $y_n^*$ for input training samples $\mathbf{x}_n$, where $\mathbf{x}_n \in \mathbf{X}\setminus \overline{\mathbf{X}}$ 
are not data centers, and $N$ is the number of data centers, $\mid\mid \cdot \mid\mid$ denotes the Frobenius norm in the multivariate case, 
and $f(\cdot)$ is the predictive function following eq~\ref{eq:pred_center}.
At each iteration we perform one gradient update step for all hyper-parameters, therefore, allowing them to be jointly learned. 

\begin{figure*}
	\centering
	\small
	\begin{tabular}{ccccc}
		{NRMSE:} & {82.99\%} & {65.32\%} & {\textbf{56.26\%}} & \\
		\includegraphics[width=0.16\textwidth]{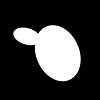} &
		\includegraphics[width=0.16\textwidth]{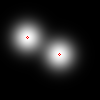} &
		\includegraphics[width=0.16\textwidth]{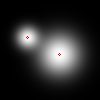} & 
		\includegraphics[width=0.16\textwidth]{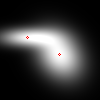} & 
		\includegraphics[width=0.23\textwidth]{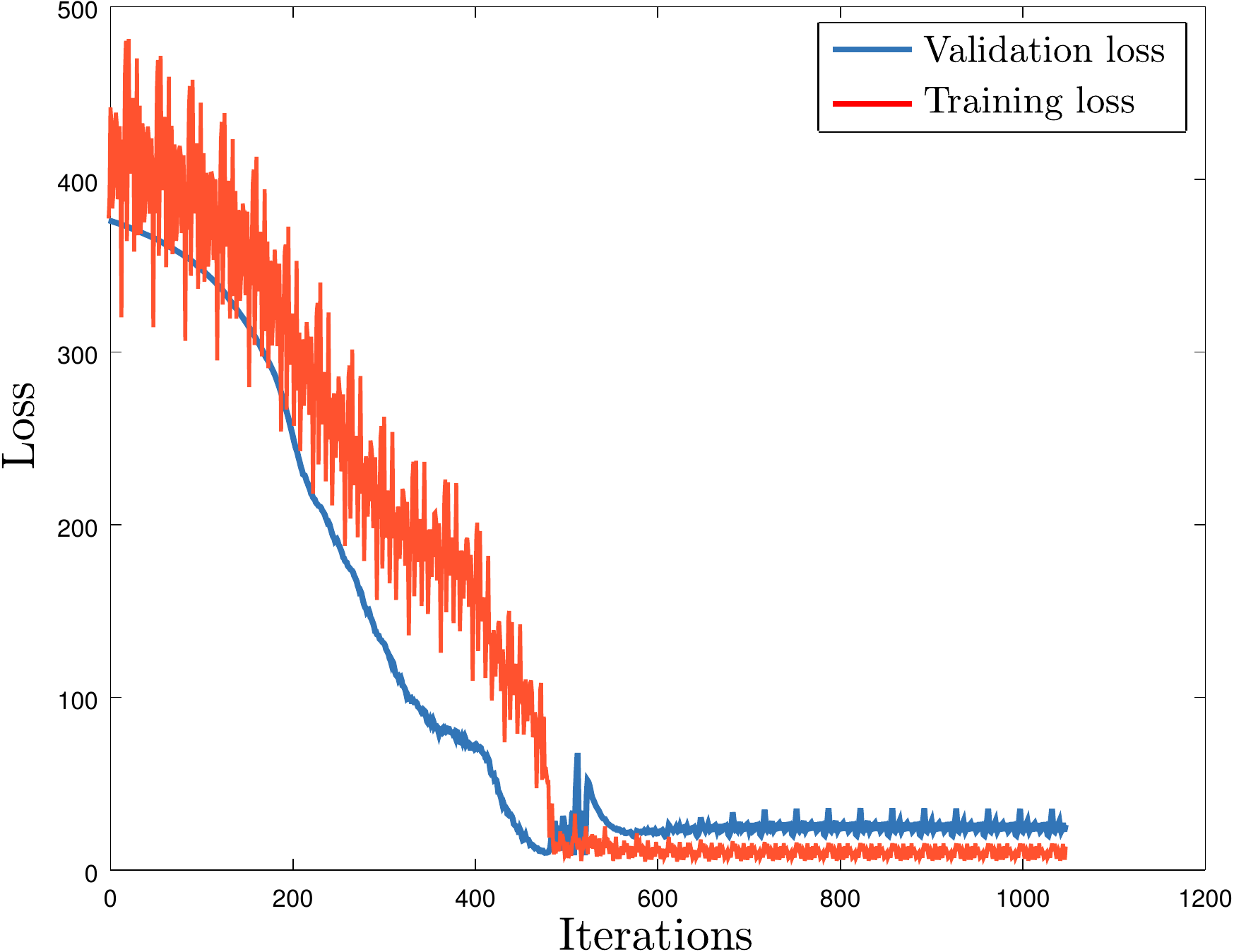} \\
		\small{Input} & \small{(i) center-GP} & \small{(ii) univariate-AGP} & \small{(iii) multivariate-AGP} & \small{(iv) losses.}\\
	\end{tabular}\\
	\caption{\small
		Pixel intensity prediction from input normalized pixel location: 
		(i) center-GP --- standard GP trained on training centers;
		(ii) univariate AGP --- proposed asymmetric model using per-center univariate lengthscale in the kernel --- eq.~\ref{eq:univariate};
		(iii) multivariate AGP --- proposed asymmetric model using per-center multivariate legnthscale --- eq.~\ref{eq:multivariate};
		(iv) losses --- training and validation losses on this data.}
	\label{fig:blobs}
\end{figure*}
\subsubsection{Univariate Multi-modal Kernel Optimization} 
\label{sssec:unikernel}
\noindent The derivative of the loss with respect to the lengthscale per center, $l_i$, for the univariate case is given by the following formulation:
\small
\begin{alignat}{1}
	\frac{\partial\mathcal{L}(f, \mathbf{y}^*)}{\partial{l_i}} &= \sum_{i=1}^N
		\left\{ \sum_{\mathbf{x}_n \in \mathbf{X} \setminus \overline{\mathbf{X}} } 2 (f(\mathbf{x}_n) - y_n^*) \right. \notag \\ &
		\left. \left[ \frac{\partial{\hat{k}}(\cdot,\mathbf{x}_n)}{\partial {l_i}} \alpha_i +  \hat{k}(\cdot,\mathbf{x}_n) 
		\left( -\hat{\mathbf{K}}^{-1} \frac{\partial{\hat{\mathbf{K}}}}{\partial{l_i}} \hat{\mathbf{K}}^{-1} \mathbf{y} \right) \right] + 2\mu l_i \right\},
	\label{eq:lossUni0}
\end{alignat}	
\begin{alignat}{3}
	\frac{\partial{\hat{k}}(\cdot,\mathbf{x}_n)}{\partial {l_i}} &=
			\left\{
                \begin{array}{ll} 
				 \frac{1}{l_i^3}(\overline{\mathbf{x}}_j - \mathbf{x}_n)(\overline{\mathbf{x}}_j - \mathbf{x}_n)^T \hat{k}(\overline{\mathbf{x}}_j,\mathbf{x}_n), 
				& \overline{\mathbf{x}}_j \in \overline{\mathbf{X}}, j = i;\\
				0, & \overline{\mathbf{x}}_j \in \overline{\mathbf{X}}, j \neq i.
				\end{array}
            \right.\\
	\frac{\partial{\hat{\mathbf{K}}}}{\partial {l_i}} &= \left( \frac{\partial{\hat{k}}(\cdot,\overline{\mathbf{x}}_m)}{\partial {l_i}} \right)_{\overline{\mathbf{x}}_m\in\overline{\mathbf{X}}},\\
	\hat{\mathbf{K}}^{-1} &= \left(\hat{k}(\overline{\mathbf{X}},\overline{\mathbf{X}}) + \sigma^2\mathbf{I}\right)^{-1},\\
	\bm{\alpha} &= \left(\hat{k}(\overline{\mathbf{X}},\overline{\mathbf{X}}) + \sigma^2 \mathbf{I}\right)^{-1} \mathbf{y},
	\label{eq:lossUni1}
\end{alignat}	
\normalsize
where we denote by $\overline{\mathbf{X}}$ the data centers, $\mathbf{\hat{K}}$ represents the asymmetric training kernel matrix, and $\mathbf{y}^*$ is a vector of training targets $y^*_n$ for training samples $\mathbf{x}_n \in \mathbf{X}\setminus \overline{\mathbf{X}}$ that are not data centers.
The lengthscale hyper-parameters, $l_i$, are estimated per training data center rather than globally.
\footnote{A simple univariate torch implementation can be found at:\\ \hspace*{10px} \emph{https://silvialaurapintea.github.io/code/gp.lua}.}
 
\subsubsection{Multivariate Multi-modal Kernel Optimization} 
\noindent In the multivariate case we learn a precision matrix, $\bm{P}_i$, rather than a scalar lengthscale per data center, $\overline{\mathbf{x}}_i \in \overline{\mathbf{X}}$. 
Therefore, we have to ensure that the precision matrix learned is symmetric. 
For this, in the gradient computation, we apply the derivations for symmetric matrices. 
\small
\begin{alignat}{1}	
	\frac{\partial\mathcal{L}(f, \mathbf{y}^*)}{\partial\bm{P}_i} &= \left[\frac{\partial\mathcal{L}(f, \mathbf{y}^*)}{\partial\bm{P}_i}\right]+ 
		\left[\frac{\partial\mathcal{L}(f,\mathbf{y}^*)}{\partial\bm{P}_i}\right]^T-\text{diag}
		\left[\frac{\partial\mathcal{L}(f,\mathbf{y}^*)}{\partial\bm{P}_i}\right]
	\label{eq:lossMulti0}
\end{alignat}
\normalsize
For gradient computation, the only change in the multivariate case is in the derivative of the kernels with respect to the per-center precision matrices, $\mathbf{P}_i$:  
\small
\begin{alignat}{3}	
	\frac{\partial{\hat{k}}(\cdot,\mathbf{x}_n)}{\partial\bm{P}_i} &= 
			\left\{
                \begin{array}{ll} 
				- \frac{1}{2}(\overline{\mathbf{x}}_j - \mathbf{x}_n)^T(\overline{\mathbf{x}}_j - \mathbf{x}_n) \hat{k}(\overline{\mathbf{x}}_j,\mathbf{x}_n), 
				& \overline{\mathbf{x}}_j \in \overline{\mathbf{X}}, j = i;\\
				0, & \overline{\mathbf{x}}_j \in \overline{\mathbf{X}}, j \neq i.
				\end{array}
            \right.\\
	\frac{\partial{\hat{\mathbf{K}}}}{\partial\bm{P}_i} &= \left( \frac{\partial{\hat{k}}(\cdot,\overline{\mathbf{x}}_m)}{\partial\bm{P}_i} \right)_{\overline{\mathbf{x}}_m\in\overline{\mathbf{X}}}.
	\label{eq:lossMulti1}
\end{alignat}
\normalsize
In this case, an additional cone projection step \cite{weinberger2009distance} is applied after each update to ensure that the precision matrix remains positive definite.
\subsection{Model Properties}
\noindent We analyze the properties considered in \cite{bellet2015metric}, from the metric learning perspective:\\	
-- \emph{Learning paradigm.} Fully supervised, as we learn from training data the best lengthscale hyper-parameters with respect to the $L_2$ prediction loss.\\
-- \emph{Form of metric.} Non-linear and local with respect to the predictive function. 
Different metrics are learned for different regions in the target space.\\
-- \emph{Scalability.} Scales with the number of data centers in the univariate case, because the kernel matrix is computed over the data centers.
Thus, is more efficient than using all training samples. 
In the multivariate case, we learn a precision matrix, and the method scales also with the number of data dimensions, making it more difficult to optimize.\\
-- \emph{Optimality of the solution.} Our learning formulation, given in equations~\ref{eq:lossUni0} and~\ref{eq:lossMulti0}, 
is a non-convex function with respect to the hyper-parameters and a global optimum is not guaranteed. 
For this reason we use an additional validation set during training on which we select among the local optima.
\section{Illustrative Results}
\label{sec:illu}
\begin{figure}
	\centering
	\includegraphics[width=0.37\textwidth]{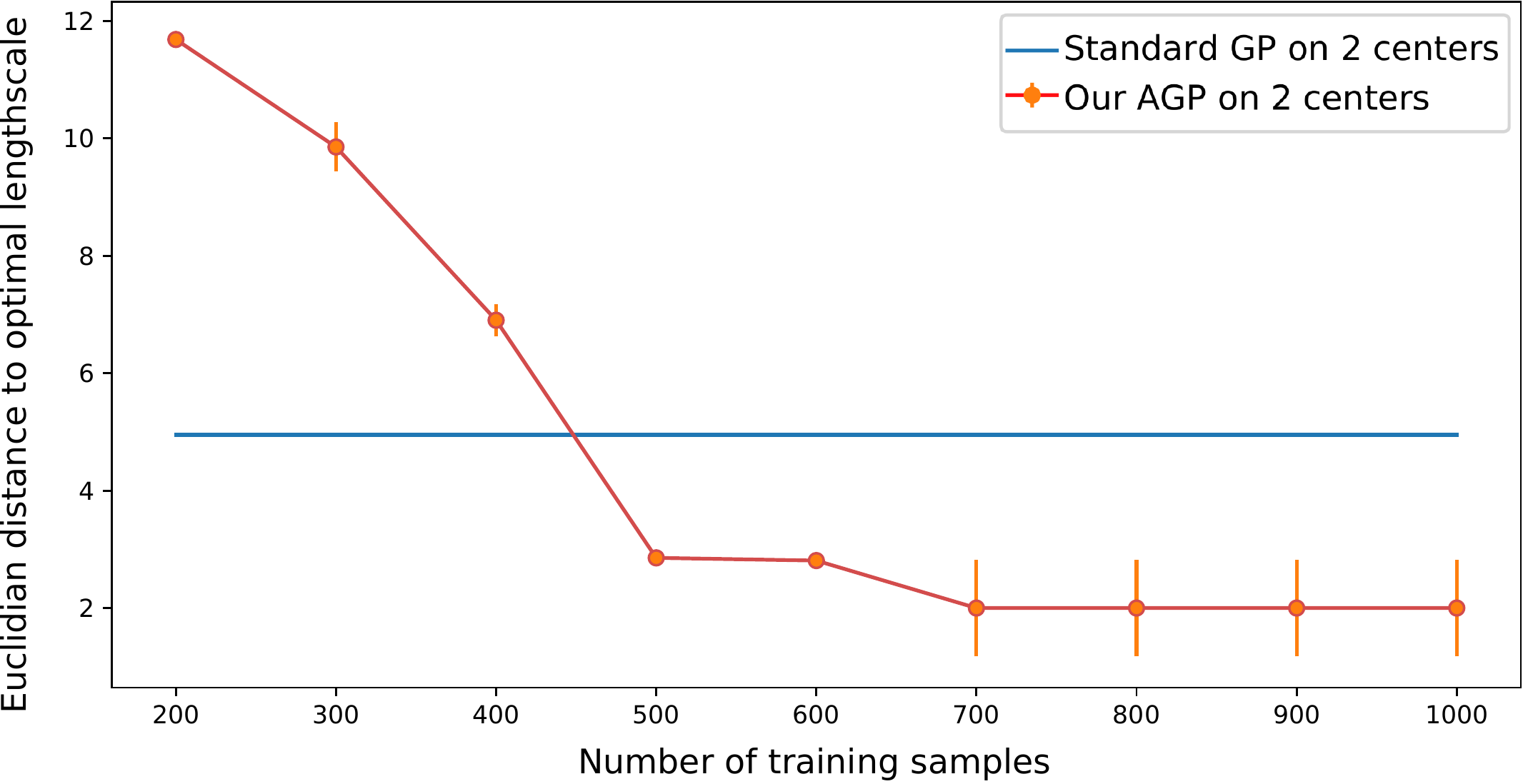} 
	\caption{\small Numeric consistency: 
		In red, the Euclidian distance between the estimated hyper-parameters by our method, using two data centers, 
		and the optimal hyper-parameter of the Gaussian Process we have sampled from.
		We plot standard deviation over 3 repetitions.
		In blue, the distance between the optimal hyper-parameter and the one of the standard GP on the two centers. 
		With more data, our estimated lengthscales approach the optimal value.
		}
	\label{fig:consistency}
\end{figure}
\noindent Figure~\ref{fig:blobs} illustrates the need for the asymmetric model. 
If all samples share the same kernel metric, figures~\ref{fig:blobs}.(ii) and~\ref{fig:blobs}.(iii) would not be possible. 
When restricting our attention to few training samples, we lose the information of how the target function varies between the samples, 
yet the per-center kernel metrics help recover this information.
In this illustration, we fix the training centers to the centers of the two ellipses. 
The $[0,1]$-normalized pixel coordinates are the input features, and the pixel intensities are the targets. 
We visualize three models: (i) \emph{center-GP} --- standard GP on the 2 data centers, and the optimal lengthscale 
hyper-parameter is found through cross validations over 100 randomly sampled training pixels; 
(ii) \emph{univariate-AGP} --- using equation~\ref{eq:univariate} with the optimal lengthscale per center estimated as in subsection~\ref{sssec:unikernel}, and 
(iii) \emph{multivariate-AGP} --- using equation~\ref{eq:multivariate}. 
In figure~\ref{fig:blobs} the univariate asymmetric model estimates better the sizes of the two blobs,
when compared to its center-based counterpart, while the multivariate asymmetric model has the lowest error, 
as it learns both the sizes and the shapes of the blobs.
We also show the training and validation losses in figure~\ref{fig:blobs}.(iv).

We additionally analyze the consistency of our approach, numerically. 
For this, we sample on a uniform grid a standard 1$D$ Gaussian Process with a fixed lengthscale set to $13.5$. 
The aim is to test if the lengthscales estimated by our method tend to the true lengthscale of the Gaussian Process from which we have sampled, as we add more training data.
We use two data centers for the kernel computation.
Figure~\ref{fig:consistency} shows in red the Euclidian distance between the two estimated hyper-parameters by our method and the optimal one.
We report standard deviation over 3 repetitions.
In blue, we show the distance between the optimal hyper-parameter and the one estimated by the standard GP through grid search, using two data centers. 
With more training samples, our estimated lengthscale parameters approach the optimal value. 
The distance does not converge to 0 as we use only two data centers in the training kernel matrix.

\section{Experiments}
\subsection{Experimental Setup}
\begin{table}[b!]
	\small
	\centering
	\caption{\small Datasets statistics.}
	\fontsize{7.2}{9.2}\selectfont
	\begin{tabular}{lrrr}
	\toprule
	Baseline & \#Trainval & \#Test & \#Features\\ \midrule	
	Temp \cite{titsias2013variational}			 & 7,117   & 3,558  & 106 \\        
	So2 \cite{titsias2013variational}			 & 15,304  & 7,652  & 27 \\            
	Airlines \cite{hoang2015unifying}			 & 2,055 K & 102 K  & 8 \\             
	UCSD \cite{chan2012counting}			     & 1,200   & 2,800  & 1,000 \\                 
	VOC-2007 \cite{chattopadhyay2017counting}    & 5,011   & 4952   & 1,000 \\ \bottomrule                
	\end{tabular}
	\label{tab:datasets}
\end{table}
\noindent \textbf{Data splitting and center selection.}
Table~\ref{tab:datasets} depicts the specifics of each dataset used.
Each dataset is split into trainval and test, following the standard way. 
Given the non-convexity of the problem, we evaluate the hyper-parameters on a small validation set after each training epoch, and keep the best.
For all datasets the validation set is obtained as 100 random samples taken from the trainval dataset. 
These samples are not used for defining the data centers or for training data statistics. 
Different center selection approaches are considered in section~\ref{ssec:center}.
We standardize the data to zero mean and unit variance per dimension, by extracting statistics over the training data excluding the validation set.\\[5px]
\noindent \textbf{Parameter setting.} In the SGD, the initial learning rate is set by looking at the plots of validation and training losses during training. 
We use a starting learning rate of $1.0e-5$ for \emph{So2}, \emph{Temp} and \emph{Airlines} datasets, 
and an initial learning rate of $1.0e-11$ for \emph{UCSD} and \emph{VOC-2007}, where the data dimensionality is considerably higher.
We use batch-SGD with mini-batches of 64 randomly selected training samples.
Given the non-convexity of the solved problem, we make use of momentum and set it to $0.9$ as advised in \cite{sutskever2013importance}.
The regularization term in the loss function, $\mu$, is set to $1.0e-5$.
For the standard GP models as well as for \emph{center-GP} --- Gaussian Process trained on data centers only --- we 
estimate the model hyper-parameters by performing grid-search and evaluating on the validation samples.
We use the same procedure for initializing our per-center lengthscales, before optimizing them in the SGD.\\[5px]
\noindent \textbf{Evaluation metric.}
For comparison with existing work we report MSE (Mean Squared Error), RMSE (Root Mean Squared Error), or NRMSE (Normalized Root Mean Squared Error) defined as:
\small
\begin{alignat}{2}
	\text{NRMSE}(\mathbf{y},\mathbf{y}^*) &= \sqrt{\frac{1}{N} \sum_n^N 
		\frac{( y_n - y^*_n)^2}{ \text{var}(\mathbf{y}_\text{train}) } },
	\label{eq:nrms}
\end{alignat}
\normalsize
where $\text{var}(\mathbf{y}_\text{train})$ is the label variance on the training data, $\mathbf{y}^*$ are the test targets, and $\mathbf{y}$ the predictions.
\begin{table}[b!]
\centering
\small
\caption{\small The effect of the center selection method when considering: K-means, random sampling, spectral clustering and GMM center definition on the \emph{Temp} dataset.}
\begin{tabular}{llllll}\toprule	
	\# Centers & \multicolumn{4}{c}{NRMSE Scores}\\ \cmidrule(r){2-5}
	           & K-Means & Sampling & Spectral & GMM\\ \midrule
	 10        & 0.602   & 0.557    & 0.579    & 0.606\\
	 50        & 0.482   & 0.467    & 0.482    & 0.523\\ \bottomrule
\end{tabular}
\label{tab:center} 	
\end{table}

\begin{table}
\small
\centering
\caption{\small Evaluation of the proposed models --- \emph{univariate-AGP} and \emph{multivariate-AGP} 
		on the large scale dataset used in \cite{hoang2015unifying}.
		Our proposed models are trained on 50 data centers. We compare our results with the methods analyzed in \cite{hoang2015unifying}: PIC, FITC, DTC. }
	\fontsize{8.2}{9.2}\selectfont
\begin{tabular}{lllll}\toprule		
		\multicolumn{2}{c}{AGP}		  & PIC							  & FITC	& DTC \\ \cmidrule(r){1-2}
		Univariate					  & Multivariate				  &         &         &     \\ \midrule
		\textbf{30.093} ($\pm$ 2.285) & \textbf{30.805} ($\pm$ 2.950) & 33.351  & 39.530  & 39.531 \\ \bottomrule
\end{tabular}
\label{tab:airlines}
\end{table}

\begin{table*}
\small
\centering
	\caption{\small
		NRMSE on the \emph{So2} and \emph{Temp} datasets for the 3 methods: 
		\emph{center-GP} --- trained on training centers only, 
		\emph{univariate-AGP} --- the univariate asymmetric model using the kernel metrics defined in eq.~\ref{eq:univariate} and, 
		\emph{multivariate-AGP} --- the multivariate asymmetric model with kernel metrics as defined in eq.~\ref{eq:multivariate}.
		Results compared with Titsias \etal \cite{titsias2013variational} and 
		\emph{GP} --- standard GP trained on randomly sampled examples. 
		We show in bold the results outperforming the baseline and underline the best result.}
\begin{tabular}{c@{\hskip 1.5in}c}
	\small
	\begin{tabular}{lll}
	\toprule	
			 & \# Samples & NRMSE \\ \midrule
	Titsias \etal 2013 & 100 & 1.004 \\ 
	GP                                           & 100 & 0.985 \\ \midrule
	center-GP                                   & 50  & \textbf{0.984} \\ 
	univariate-AGP                               & 50  & \textbf{0.846} \\ 
	multivariate-AGP                             & 50  & \textbf{0.863} \\ \midrule
	center-GP                                   & 10  & 0.985 \\ 
	univariate-AGP                               & 10  & \textbf{0.818} \\ 
	multivariate-AGP                             & 10  & \textbf{\underline{0.808}} \\ \bottomrule
	\end{tabular} &
	\small
	\begin{tabular}{lll}
	\toprule	
			 & \# Samples & NRMSE \\ \midrule
	Titsias \etal 2013 & 100 & 0.489 \\ 
	GP                                           & 100 & 0.533 \\ \midrule
	center-GP                                   & 50  & 0.559 \\ 
	univariate-AGP                               & 50  & \textbf{0.482} \\ 
	multivariate-AGP                             & 50  & \textbf{\underline{0.445}} \\ \midrule
	center-GP                                   & 10  & 0.642 \\ 
	univariate-AGP                               & 10  & 0.602 \\ 
	multivariate-AGP                             & 10  & 0.493 \\ \bottomrule
	\end{tabular} \\
	(a) \emph{So2} data evaluation. & (b) \emph{Temp} data evaluation.\\
	\end{tabular}
\label{tab:res2} 	
\end{table*}

\subsection{Center Selection}
\label{ssec:center}
\noindent Here we analyze the effect of the center selection method on the overall performance of our method.
For this we use the \emph{Temp} dataset which has $106$ dimensions per sample.
We consider four center selection methods: K-Means, random sampling, spectral clustering and GMM (Guassian Mixture Model).
We test on our \emph{univariate-AGP} --- using univariate individualized lengthscales --- 
with 10 and 50 centers, respectively.	

Table \ref{tab:center} indicates that the choice of the centers is not essential as all methods perform similar.
Given that the strength of the model is in learning individualized hyper-parameters and less in the method used for defining centers,
in our subsequent experiments we rely on k-means clustering.

\subsection{Multi-modal Approach Evaluation} 
\noindent Table~\ref{tab:res2} depicts the results of our approaches on the \emph{So2} and \emph{Temp} datasets 
when compared to \cite{titsias2013variational} and with the standard Gaussian Process model. 
The gain brought by the multi-modal asymmetric methods over the center-based Gaussian Process and the standard 
Gaussian Process is more obvious for the \emph{Temp} dataset.
This can be explained by the larger number of dimensions to learn from, in the multivariate case. 
On both datasets, the proposed models outperform \cite{titsias2013variational}, while using only 50 data centers. 

The performance decreases slightly with the increase in the number of data centers for the multivariate asymmetric models.
This is due to the model being trained for the same number of iterations as the univariate case, while having to learn a larger number of parameters --- 
a precision matrix of size $D \times D$. 
With the increase in data dimensionality, the multivariate model becomes considerably slower and harder to optimize.
This represents a drawback of the proposed multivariate approach. 
The variability in the target space also affects the ease with which a good solution is found. 

\begin{table}
	\small
	\caption{\small RMSE results on the VOC-2007 general object counting dataset. 
	The first two ``glance" models use global image features learned in a deep learning framework, which is similar to us. 
	The last two models use local information by dividing the image into a 3$\times$3 grid and extracting deep learning features from each cell. 
	Our method outperforms the models relying on global image features.}
	\begin{tabular}{ll}
	\toprule		
	\cite{chattopadhyay2017counting} glance-noft-2L			   & 0.50 ($\pm$ 0.02) \\
	\cite{chattopadhyay2017counting} glance-sos-2L			   & 0.51 ($\pm$ 0.02) \\ \midrule
	\cite{chattopadhyay2017counting} aso-sub-ft-1L-3$\times$3L & 0.43 ($\pm$ 0.01) \\ 
	\cite{chattopadhyay2017counting} seq-sub-ft-3$\times$3	   & 0.42 ($\pm$ 0.01) \\ \midrule
	AGP-25 Univariate										   & 0.43 ($\pm$ 0.002)\\ \bottomrule
	\end{tabular}
	\label{tab:voc}
\end{table}
\subsection{State-of-the-art Comparison}
\subsubsection{Large Scale Regression Problem}
\noindent Here, we consider a more challenging task where the number of training samples is markedly high.
We compare against effective state-of-the-art methods that focus on the same problem as we do --- representing the data using an 
informative subset while retaining the descriptiveness of the model \cite{hoang2015unifying}.  
We use the \emph{Airlines} dataset of \cite{hoang2015unifying} containing over $2,000,000$ training samples. 
The models considered are: DTC (Deterministic Training Conditional) \cite{seeger2003fast}, 
FITC (Fully Independent Training Conditional) \cite{snelson2005sparse},
and PIC (Partially Independent Conditional) \cite{snelson2007local}.
For evaluating our models we repeat each experiment three times and report the mean RMSE (Root Mean Squared Error) together with the standard deviation.
Table \ref{tab:airlines} shows that our proposed \emph{AGP} models perform well when dealing with a prohibitive number of training samples.
Our approach outperforms existing methods \cite{seeger2003fast, snelson2005sparse, snelson2007local} while using only a limited number of data centers.
\subsubsection{Realistic Data: Counting from Images}
\begin{figure}[b!]
	\small
	\centering
	\begin{tabular}{c}
		(a) Squared losses on UCSD.\\
		\includegraphics[width=0.35\textwidth]{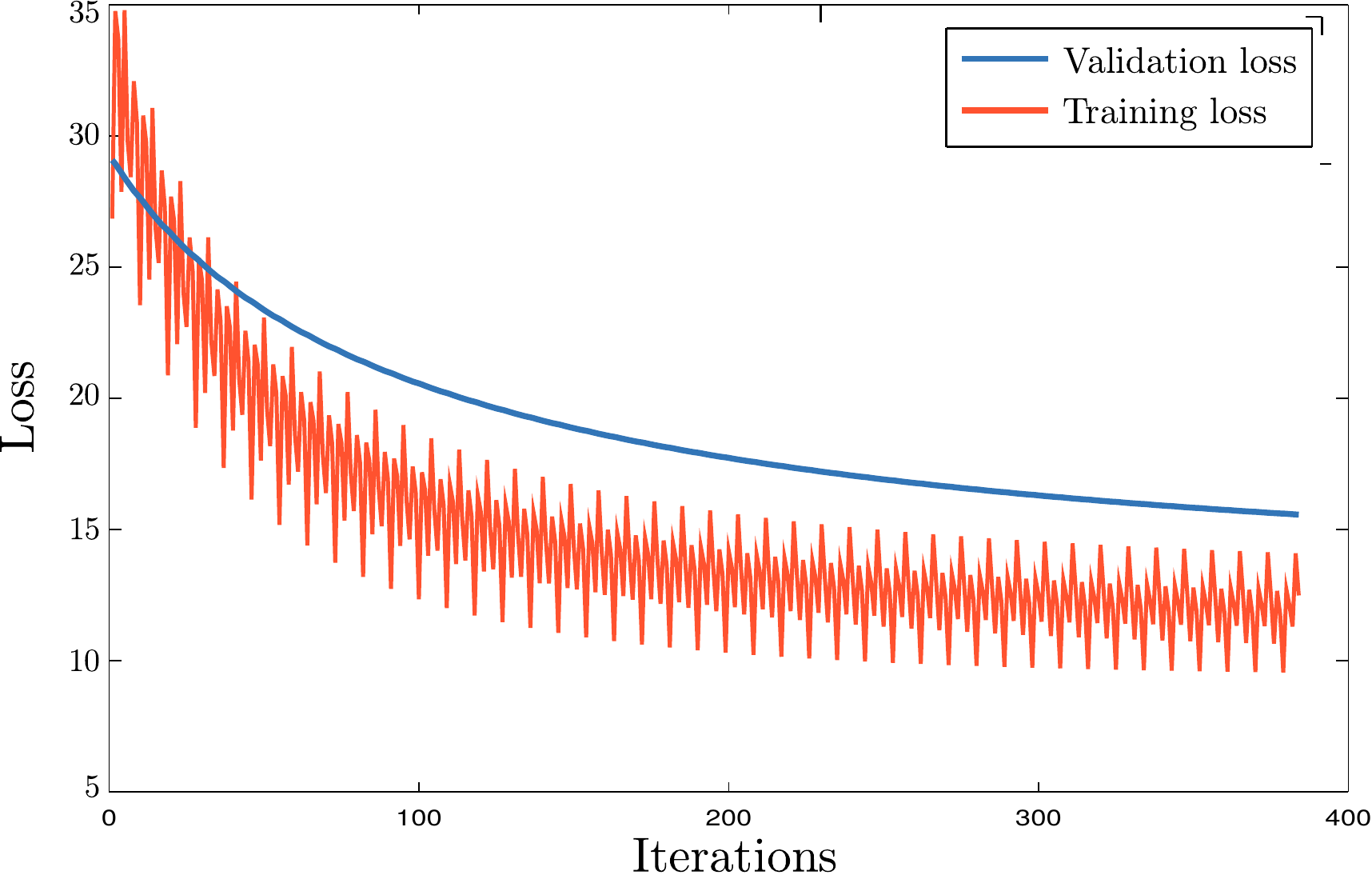} \\[5px]
		(b) MSE on UCSD. \\ 
		\begin{tabular}{ll}
			\toprule		
			\cite{dalal2005histograms}              & 39.75 \\
			\cite{felzenszwalb2008discriminatively} & 24.72 \\
			\cite{chan2012counting} 			    & 9.95 \\ \midrule
			AGP-10 Univariate & 16.37 ($\pm$ 0.41)\\
			AGP-25 Univariate & 13.90 ($\pm$ 0.05)\\ \bottomrule
		\end{tabular} \\
	\end{tabular}
	\caption{\small (a) MSE results on the UCSD pedestrians counting dataset when compared with three prior works \cite{dalal2005histograms,felzenszwalb2008discriminatively,chan2012counting}. 
	We obtain comparable performance with prior work, though we use only global deep learning features, while \cite{chan2012counting} relies on motion segmentation masks. 
	(b) The training and validation squared losses on the UCSD pedestrian counting dataset.}
\label{fig:ucsd}
\end{figure}
\noindent We test our regression approach on two realistic image datasets --- UCSD \cite{chan2012counting} and VOC-2007 \cite{pascal-voc-2007} --- 
for people and generic object counting, respectively. 
Given the image data, we rely on deep learning features. 
We extract 1,000 dimensional features as the output of the fully-connected layer of the ResNet-50 \cite{he2016deep} pretrained on ImageNet \cite{russakovsky2015imagenet}.  
For computational efficiency here we use only the univariate version of our approach with 25 centers, since the multivariate version requires optimizing a precision matrix of size $1,000 \times 1,000$. 

\vspace{5px} \noindent \textbf{Pedestrian counting.}
Figure~\ref{fig:ucsd} shows the results of our approach on the realistic problem of pedestrian counting from images on the UCSD dataset, together with the training and validation losses. 
We compare with \cite{dalal2005histograms} that uses low level image features, \cite{felzenszwalb2008discriminatively} relying on a person detection method specifically trained for the task,
and \cite{chan2012counting} which employs motion segmentation masks. 
Unlike these methods, we do not use either motion segmentation masks or class specific detectors.
We use only global image features extracted from a pretrained deep network, and we manage to obtain comparable performance with \cite{chan2012counting}, 
while greatly outperforming \cite{dalal2005histograms, felzenszwalb2008discriminatively}.  

\vspace{5px}\noindent \textbf{Generic object counting.}
In table~\ref{tab:voc} the goal is generic object counting on the VOC-2007 generic object dataset.
We compare with the set of models proposed in the very recent deep learning method of \cite{chattopadhyay2017counting}. 
Our features are extracted from a pretrained deep learning model, while theirs are specifically fine-tuned for this counting task. 
We outperform their \emph{``glance"} models, which similar to us, rely on global image features. 
We additionally obtain comparable performance to \emph{aso-sub-ft-1L-3$\times$3} and \emph{seq-sub-ft-3$\times$3} which rely on local image information, 
as they divide each image into a 3$\times$3 grid and extract features from each cell. 
It is worthwhile noting that we use only 25 data centers for computing the kernel matrix, and achieve comparable performance with methods relying on stronger features.
These results support our approach. 
\section{Conclusions}
\noindent This work brings forth an asymmetric kernel for the Gaussian Process model. 
This encompasses three components:
(i) training on training centers only, 
(ii) learning individualized kernel metrics per center and, 
(iii) extending the lengthscale hyper-parameter to a precision matrix, thus learning not only the appropriate size but also the shape in the kernel metric. 
Due to the limitations imposed by the dependency between per-center hyper-parameters, we discriminatively solve the problem through metric learning.
Individualized kernel metrics entail the loss of the symmetry in the kernel matrix.
However, this has the gain of better describing the target function in the neighborhood of the center points, used for kernel computation.

{
\small
\vspace{5px}\noindent \textbf{Acknowledgements} Research supported by the Dutch national program COMMIT.
Thanks to dr. Mijung Park and dr. Marco Loog for the insightful suggestions and advice.
}
{
	\small
	\bibliographystyle{model2-names} 
	\bibliography{agp}
}
\end{document}